\documentclass[sigplan,screen]{acmart}

\AtBeginDocument{%
  }

\copyrightyear{2025}
\acmYear{2025}
\setcopyright{cc}
\setcctype{by}
\acmConference[UCC '25]{2025 IEEE/ACM 18th International Conference on Utility and Cloud Computing}{December 1--4, 2025}{Nantes, France}
\acmBooktitle{2025 IEEE/ACM 18th International Conference on Utility and Cloud Computing (UCC '25), December 1--4, 2025, Nantes, France}\acmDOI{10.1145/3773274.3774931}
\acmISBN{979-8-4007-2285-1/2025/12}

\usepackage{algpseudocode}
\usepackage{placeins}
\usepackage{booktabs}
\usepackage{float}
\usepackage{caption}
\usepackage{cuted}

\begin{document}

\title[HybridVFL]{HybridVFL: Disentangled Feature Learning for Edge-Enabled Vertical Federated Multimodal Classification}

\author{Mostafa Anoosha}
\email{m.anoosha@hull.ac.uk}
\affiliation{%
  \department{School of Digital and Physical Science}
  \institution{University of Hull}
  \city{Hull}
  \country{UK}
}

\author{Zeinab Dehghani}
\email{z.dehghani@hull.ac.uk}
\affiliation{%
  \department{School of Digital and Physical Science}
  \institution{University of Hull}
  \city{Hull}
  \country{UK}
}

\author{Kuniko Paxton}
\email{k.azuma@hull.ac.uk}
\affiliation{%
  \department{School of Digital and Physical Science}
  \institution{University of Hull}
  \city{Hull}
  \country{UK}
}

\author{Koorosh Aslansefat}
\email{k.aslansefat@hull.ac.uk}
\affiliation{%
  \department{School of Digital and Physical Science}
  \institution{University of Hull}
  \city{Hull}
  \country{UK}
}

\author{Dhaval Thakker}
\email{d.thakker@hull.ac.uk}
\affiliation{%
  \department{School of Digital and Physical Science}
  \institution{University of Hull}
  \city{Hull}
  \country{UK}
}

\renewcommand{\shortauthors}{Anoosha et al.}

\begin{abstract}
  Vertical Federated Learning (VFL) offers a privacy-preserving paradigm for Edge AI scenarios like mobile health diagnostics, where sensitive multimodal data reside on distributed, resource-constrained devices. Yet, standard VFL systems often suffer performance limitations due to simplistic feature fusion. This paper introduces HybridVFL, a novel framework designed to overcome this bottleneck by employing client-side feature disentanglement paired with a server-side cross-modal transformer for context-aware fusion. Through systematic evaluation on the multimodal HAM10000 skin lesion dataset, we demonstrate that HybridVFL significantly outperforms standard federated baselines, validating the criticality of advanced fusion mechanisms in robust, privacy-preserving systems.
\end{abstract}



\maketitle

\section{Introduction}

    Since Google introduced the concept of federated learning (FL)\cite{mcmahan2017communication}, distributed learning has gained traction in fields where centralizing personal data is difficult, such as the healthcare sector operating under stringent data privacy regulations like the European General Data Protection Regulation\cite{regulation2018general}. While Horizontal FL (HFL) requires uniform feature spaces across clients~\cite{yang2019federated}, real-world collaborations often involve diverse organizations holding different features for overlapping user groups~\cite{yang2023survey}. Vertical Federated Learning (VFL) addresses this by enabling collaboration when data sets share the same sample ID space but differ in feature space~\cite{yang2019federated}. Technically, this approach is simultaneously expected to enhance model expressiveness by combining multiple and rich features \cite{che2023multimodal}. Despite the high potential of VFL applications in real-world scenarios, prior research has predominantly focused on HFL, with VFL receiving relatively limited attention \cite{khan2025vertical}. 
    
    A key aspect of VFL is multimodal processing on the server side. This can be broadly categorized into two types: (1) between homogeneous data, which is the same data structures, such as between images under different protocols \cite{yan2024cross}, and (2) heterogeneous data, such as \cite{mandal2024distributed} of images and tabular data. In the latter case, data modality handling itself is complex due to differences in data structure and label asymmetry; only the tabular side holds labels, which occurs, increasing the learning difficulty. Medical data typically coexists as images and electronic health records (EHR), making this need particularly high \cite{rauniyar2023federated,qayyum2022collaborative,guan2024federated}. While reports indicate that incorporating EHR in centralized approaches improves classification performance compared to inputting images alone \cite{ou2022deep}, most FL evaluations remain confined to the HFL framework \cite{dai2024federated,agbley2021multimodal,ouyang2023harmony,yuan2024communication}, and the effects of heterogeneous feature space collaboration remain insufficiently evaluated. Furthermore, even in studies considering image and tabular pairs within VFL, the feature handling tends to remain simple concatenation \cite{sundar2024toward}, and comparative analyses of effects between modal fusion strategies are lacking. Although disentangled cross-modal fusion \cite{yan2024cross} is a method that splits into modality-invariant and -specific, and then fuses to combine heterogeneous information more effectively and semantically, the prior research has not systematically explored this method for VFL with image and tabular.
    
    To bridge this gap, this study focuses on the medical domain, which is a high-demand field of FVL \cite{pfitzner2021federated}, and systematically compares (1) concatenation, (2) simple cross-attention fusion, and (3) disentangled cross-modal fusion under the VFL environment, analyzing how fusion design differences impact performance. Evaluation utilizes dermatological lesion images and clinical metadata from Human Against Machine with 10000 (HAM10000) \cite{tschandl2018ham10000}. Dermatological lesion classification is a representative task well-suited for mobile health and clinical edge computing \cite{alasbali2025privacy}, making it highly significant for VFL implementation. Our study suggests that semantic fusion is more promising than simple concatenation or simple cross-modal attention, further leveraging the VFL advantage of utilizing heterogeneous features in distributed environments. The three pillars of VFL are robust multimodal fusion design, privacy protection, and fairness. We eventually aim to develop \textbf{HybridVFL} that integrates these elements. This paper focuses specifically on examining multimodal fusion design as the first step.

    \subsection{Main Contribution}
    
        Considering existent gaps in federated learning on truly diverse data modalities and simplistic fusion, we assess the effectiveness of disentangled cross-modal fusion and compare it with other approaches for dermatological classification in HybridVFL across heterogeneous image and tabular feature spaces. This distils practical guidance on fusion selection for the HybridVFL dermatological model.    

\section{Related Work}

    This work builds on three foundational pillars: \textit{Vertical Federated Learning (VFL)}, \textit{Multimodal Integration}, and \textit{Disentangled Cross-Modal Fusion}. VFL enables collaborative model training across organisations with complementary features, while Multimodal Fusion integrates diverse modalities into unified, privacy-preserving representations. Disentangled Cross-Modal Fusion further refines this process by separating representations and combining them via attention. Together, these advances lay the foundation for our proposed \textit{HybridVFL} framework for scalable, privacy-aware cross-institutional learning.

    \subsection{Vertical Federated Learning (VFL)}
    
        Federated Learning (FL) is generally categorised into horizontal and vertical paradigms~\cite{yang2019federated}. In \textit{HFL}, clients share the same feature space but hold different sample subsets, while \textit{VFL} supports collaboration among institutions with distinct feature spaces for overlapping entities~\cite{kairouz2021advances}. Each participant exchanges encrypted intermediate representations, such as embeddings or gradients, securely aggregated by a central server~\cite{hardy2017private,cheng2021secureboost}.
        
        Modern VFL frameworks have evolved from basic cryptographic logistic regression~\cite{hardy2017private} toward more practical systems like \textit{SecureBoost}~\cite{cheng2021secureboost} for tree-based learning and \textit{Falcon}~\cite{wu2023falcon} for interpretable deep learning.
        
        Recent frameworks such as \textit{Falcon}~\cite{wu2023falcon} balance scalability and interpretability by combining threshold homomorphic encryption with feature-importance analysis. Overall, VFL has evolved from cryptography-based designs~\cite{hardy2017private,yang2019federated} toward more practical and interpretable systems~\cite{cheng2021secureboost}, motivating our proposed \textit{HybridVFL} framework for advancing multimodality, scalability, and trustworthy AI in cross-institutional learning.

    \subsection{Multimodal Integration in Federated Learning}
    
        Leveraging complementary data modalities has become central to FL, particularly in VFL, where institutions hold distinct modalities that must be integrated without exposing raw data. Surveys such as Che et al.~\cite{che2023multimodal} and Baltrušaitis et al.~\cite{baltruvsaitis2018multimodal} summarise progress in model architectures and optimisation strategies addressing privacy, heterogeneity, and interpretability. Transformer-based designs, inspired by Vaswani et al.~\cite{vaswani2017attention} and extended to multimodal contexts by Tsai et al.~\cite{tsai2019multimodal}, have proven highly effective for cross-modal representation learning. In the medical domain, such approaches are increasingly applied to federated multimodal diagnosis tasks, with benchmarks like HAM10000 supporting evaluation~\cite{tschandl2018ham10000}.

    \subsection{Disentangled Cross-Modal Fusion}
    
        Building on multimodal FL, recent works have adopted disentangled representation learning combined with attention-based and Transformer-style architectures~\cite{vaswani2017attention,tsai2019multimodal} to enhance alignment and interpretability in VFL systems. Yan et al.~\cite{yan2024cross} introduced \textit{Fed-CRFD}, which separates invariant and modality-specific embeddings with cross-client consistency, forming the conceptual basis for our disentanglement strategy. These methods broadly aim to isolate shared from modality-specific information to enable context-aware representation learning~\cite{higgins2018towards,locatello2019challenging}.  
        \sloppy
        In parallel, contrastive alignment strategies such as \textit{MOON}~\cite{li2021model} demonstrate the effectiveness of enforcing feature consistency across federated clients. Together, these advances highlight the potential of disentangled and contrastive fusion to create robust, interpretable multimodal VFL systems.

\section{The HybridVFL Architectural Framework}

    \label{sec:method}
    The proposed \textbf{HybridVFL} framework mitigates the fusion bottleneck in Vertical Federated Learning (VFL)~\cite{yang2019federated,li2020federated} 
    by combining client-side feature disentanglement~\cite{higgins2018towards,locatello2019challenging} 
    with a server-side contextual fusion module based on Transformer attention~\cite{jin2021cafe}. 
    The data flow proceeds from raw, siloed client data to disentangled embeddings and a unified global prediction.

    \subsection{System Overview}

        HybridVFL operates in a standard VFL configuration~\cite{cheng2021secureboost} 
        with two non-colluding clients and a semi-honest central server.
        Formally, the shared dataset is defined in Eq.~\ref{eq:dataset}:
        \begin{equation}
        \mathcal{D}=\{(x_I^{(i)},x_T^{(i)},y^{(i)})\}_{i=1}^{N},
        \label{eq:dataset}
        \end{equation}
        where $x_I^{(i)}\!\in\!\mathbb{R}^{H\times W\times C}$ are dermoscopic images owned by the \textit{Image Client} $\mathcal{C}_I$,
        $x_T^{(i)}\!\in\!\mathbb{R}^{p}$ are structured metadata (e.g., age, gender, lesion site) held by the \textit{Tabular Client} $\mathcal{C}_T$, and
        $y^{(i)}\!\in\!\{0,1\}^K$ are one-hot encoded diagnostic labels across $K=7$ lesion types, maintained only by the server $\mathcal{S}$. 
        Raw data are never exchanged, ensuring compliance with privacy regulations such as HIPAA and GDPR~\cite{regulation2018general}.

    \subsection{Client-Side Disentangled Encoding}
    
\begin{figure}[t]
        \centering
        \includegraphics[width=0.8\linewidth]{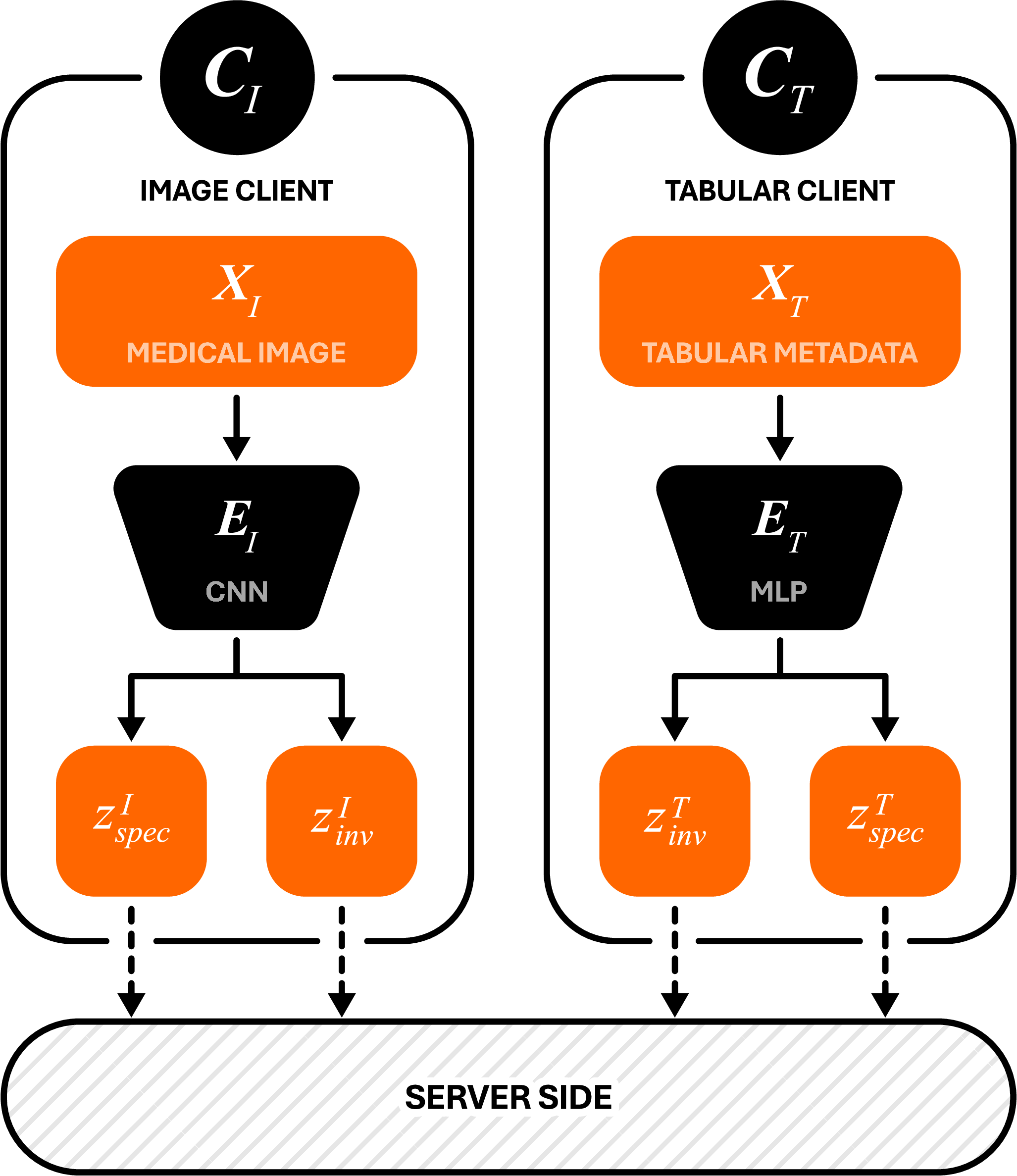}
        \caption{Client-side architecture: Specialized encoders ($E_I, E_T$) disentangle raw data into modality-specific ($z_{spec}$) and invariant ($z_{inv}$) embeddings for transmission to the server.}
        \Description{Diagram showing two clients, one with image data and one with tabular data, each producing invariant and specific embeddings that are sent to a central server.}
        \label{fig:placeholder}
    \end{figure}

        To enable feature-level collaboration without sharing raw data, each client employs a dual-output encoder that decomposes its input into invariant and modality-specific embeddings.  
        The encoder operations are formally expressed in Eq.~\ref{eq:encoders}:
        \begin{equation}
        (z_{\text{inv}}^I,z_{\text{spec}}^I)=E_I(x_I;\theta_I), \qquad
        (z_{\text{inv}}^T,z_{\text{spec}}^T)=E_T(x_T;\theta_T),
        \label{eq:encoders}
        \end{equation}
        where $E_I$ (a CNN) and $E_T$ (an MLP) are modality-specific encoders~\cite{higgins2018towards,locatello2019challenging,tsai2019multimodal}. 
        As shown in Fig.~\ref{fig:placeholder}, the \textit{Image Client} ($\mathcal{C}_I$) and \textit{Tabular Client} ($\mathcal{C}_T$) process their raw data separately and send only their disentangled embeddings to the server.
        Here, $z_{\text{inv}}^\bullet$ captures shared semantics relevant to malignancy, while $z_{\text{spec}}^\bullet$ retains modality-specific information. 
        Only these embeddings are transmitted to the server; raw features remain local.

    \subsection{Server-Side Disentangled Fusion}
    
\begin{figure}[t]
            \centering
            \includegraphics[width=0.8\linewidth]{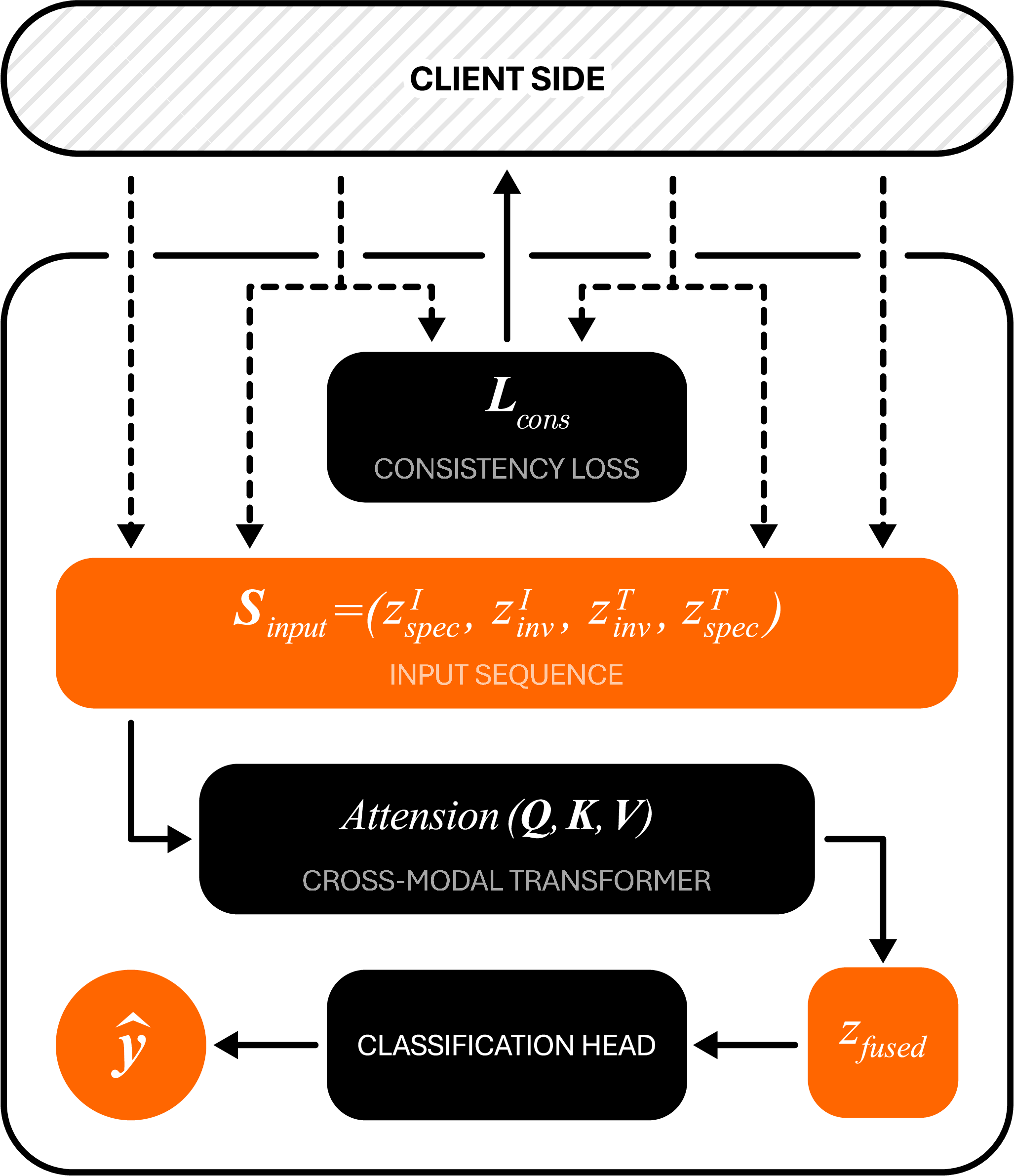}
            \caption{The server receives the four embeddings from the clients. 
            First, a consistency loss ($\mathcal{L}_{cons}$) is applied to align the invariant features. 
            Next, all four embeddings are combined into an input sequence ($S_{input}$) and fed into a Cross-Modal Transformer. 
            The transformer applies self-attention to produce a single fused representation ($z_{fused}$), which is then passed to a classification head to generate the final prediction ($\hat{y}$).}
            \Description{Diagram showing the server aligning invariant embeddings with a consistency loss, fusing four embeddings using a Cross-Modal Transformer, and producing a final prediction through a classification head.}
            \label{fig:placeholder1}
        \end{figure}

        As shown in Fig.~\ref{fig:placeholder1}, the server receives the invariant and specific embeddings from both clients and performs two key operations. 
        
        \textbf{Stage 1: Invariant Alignment.}
        Before fusing representations, the server enforces alignment between invariant embeddings from both clients by minimising the cosine-based consistency loss defined in Eq.~\ref{eq:consistency}:
        \begin{equation}
        \mathcal{L}_{\text{cons}} = 1 -
        \frac{\langle z_{\text{inv}}^I, z_{\text{inv}}^T \rangle}
        {\|z_{\text{inv}}^I\|_2 \|z_{\text{inv}}^T\|_2}.
        \label{eq:consistency}
        \end{equation}
        This regulariser promotes a shared latent representation across modalities, improving generalisation~\cite{li2021model}.
        
        \textbf{Stage 2: Contextual Fusion.}
        After alignment, all embeddings are concatenated into a single token sequence as shown in Eq.~\ref{eq:sequence}:
        \begin{equation}
        S=[z_{\text{inv}}^I,\,z_{\text{spec}}^I,\,z_{\text{inv}}^T,\,z_{\text{spec}}^T],
        \label{eq:sequence}
        \end{equation}
        which is then processed by a Transformer encoder $F_\phi(\cdot)$ that models inter-modality dependencies via multi-head self-attention, defined in Eq.~\ref{eq:attention}:
        \begin{equation}
        \text{Attention}(Q,K,V)=\text{softmax}\!\left(\tfrac{QK^\top}{\sqrt{d_k}}\right)V.
        \label{eq:attention}
        \end{equation}
        Finally, the contextualised representations from the Transformer are aggregated through mean pooling as expressed in Eq.~\ref{eq:zfused}:
        \begin{equation}
        z_{\text{fused}}=\text{Pool}(F_\phi(S)),
        \label{eq:zfused}
        \end{equation}
        Where the pooling operation follows the convention used in multimodal Transformers~\cite{tsai2019multimodal,baltruvsaitis2018multimodal}.

    \subsection{Global Training and Optimisation}

        The fused embedding $z_{\text{fused}}$ is passed through a lightweight classifier that outputs the predicted class probabilities via a softmax activation, as shown in Eq.~\ref{eq:prediction}:
        \begin{equation}
        \hat{y}=\text{softmax}(h_\psi(z_{\text{fused}})),
        \label{eq:prediction}
        \end{equation}
        where $\hat{y} \in \mathbb{R}^K$ represents the probability distribution over the $K=7$ lesion classes.
        
        HybridVFL is trained end-to-end by minimising a composite loss combining the categorical cross-entropy objective and the alignment regulariser from Eq.~\ref{eq:consistency}. The overall training objective is defined in Eq.~\ref{eq:total_loss}:
        \begin{equation}
        \mathcal{L}_{\text{total}} = -\frac{1}{N}\sum_{i=1}^{N} \sum_{k=1}^{K} y_k^{(i)} \log \hat{y}_k^{(i)} + \lambda_{\text{cons}} \, \mathcal{L}_{\text{cons}}.
        \label{eq:total_loss}
        \end{equation}
        The coefficient $\lambda_{\text{cons}}$ balances the contribution of invariant-space alignment to the main classification objective.

    \subsection{Communication Efficiency}
    
        Unlike centralized training, HybridVFL requires exchanging embeddings between clients and the server at every training round. For our architecture (1600 total embedding dimensions), this incurs approximately 6.4 KB per sample per round—significantly lower than transmitting raw medical images (~120 KB for $100\times100$ float32 inputs). This low bandwidth requirement makes HybridVFL highly viable for resource-constrained edge environments.

\section{Numerical Results}

    This section presents a comprehensive empirical evaluation of the proposed \textbf{HybridVFL} framework. 
    The experiments are designed to systematically validate the effectiveness of the disentangled cross-modal fusion architecture by comparing it against both centralised and federated baselines. 
    The analysis quantifies the performance gains achieved by the proposed approach and contextualises its performance within the inherent trade-offs of the Vertical Federated Learning (VFL) paradigm. 
    The evaluation demonstrates not only the superiority of \textbf{HybridVFL} in a privacy-preserving setting but also its ability to achieve performance levels competitive with those of a centralised model.

    \subsection{Dataset and Preprocessing}
    
        All experiments use the HAM10000 (\textit{Human Against Machine with 10000}) dataset~\cite{tschandl2018ham10000}, a standard benchmark for dermatological lesion classification. 
        It contains 10{,}015 dermoscopic images with metadata including patient age, gender, and lesion site, providing a realistic multimodal dataset for VFL. 
        The \textit{Image Client} ($\mathcal{C}_I$) holds image data, the \textit{Tabular Client} ($\mathcal{C}_T$) stores the corresponding EHR records, and the server maintains diagnostic labels ($y$). 
        The task is a multi-class classification across the seven verified skin lesion types, with data vertically partitioned as described in Section~\ref{sec:method}.

    \subsection{Evaluation Metrics}
    
        Model performance is assessed using standard classification metrics:
        \begin{itemize}
            \item \textbf{Macro F1-Score, Precision, \& Recall:} Unweighted arithmetic means across all seven classes. These metrics ensure balanced performance evaluation, preventing domination by frequent classes and accurately reflecting detection capability for rare lesions.
            \item \textbf{Test Accuracy:} The overall proportion of correctly classified samples.
            \item \textbf{Balanced Accuracy:} The arithmetic mean of recall across all seven classes. This provides a more reliable performance measure than standard accuracy in the presence of the severe class imbalance typical of HAM10000.
        \end{itemize}

    \subsection{Comparative Models}
    
        The \textbf{HybridVFL} framework is compared against several representative models covering distinct architectural paradigms and data-access settings. 
        Together, they define both a practical federated baseline and a centralised upper-bound oracle.
        
        \begin{enumerate}
            \item \textbf{Centralized Image Only:} 
            A privacy-free baseline using only image data processed by a CNN, establishing unimodal performance and enabling assessment of the added value of tabular metadata.
            
            \item \textbf{Centralized Multimodal:} 
            A centralised oracle combining CNN and MLP features through naive concatenation, representing the upper bound achievable with full data access.
            
            \item \textbf{Simple Concatenation VFL:} 
            A standard federated baseline where clients send local embeddings to the server, which concatenates them for classification, mirroring early VFL frameworks.
            
            \item \textbf{HybridVFL (Proposed):} 
            The proposed framework (Section~\ref{sec:method}) integrates disentangled client encoders ($E_I, E_T$), a cosine-based consistency loss ($\mathcal{L}_{\text{cons}}$) for invariant alignment, and a Cross-Modal Transformer ($F_{\phi}$) for contextual multimodal fusion.
        \end{enumerate}

    \subsection{Comparative Performance Analysis}
    
        The core quantitative results of this study are summarised in Table~\ref{tab:performance}, which provides a direct comparison between the proposed \textbf{HybridVFL} framework and the established baseline models. 
        The subsequent analysis examines these results in detail to validate the study's primary hypotheses and to assess the effectiveness of the disentangled cross-modal fusion strategy under the Vertical Federated Learning (VFL) paradigm.
        
        \begin{strip}
        \centering
        \small  
        \captionof{table}{Comparative Performance on the HAM10000 Dataset. The best federated result is highlighted in bold.}
        \label{tab:performance}
        \begin{tabular*}{\textwidth}{@{\extracolsep{\fill}} l c c c c c}
        \toprule
        \textbf{Model Architecture} & \textbf{Macro F1} & \textbf{Macro Precision} & \textbf{Macro Recall} & \multicolumn{2}{c}{\textbf{Accuracy}} \\
        \cline{5-6}
         &  &  &  & \textbf{Test} & \textbf{Balanced} \\
        \midrule
        Centralized Image-Only & 0.8771 $\pm$ 0.0010 & 0.8538 $\pm$ 0.0008 & 0.9074 $\pm$ 0.0018 & 0.8515 $\pm$ 0.0082 & 0.9074 $\pm$ 0.0093 \\
        Centralized Multimodal & 0.9222 $\pm$ 0.0024 & 0.9130 $\pm$ 0.0027 & 0.9325 $\pm$ 0.0081 & 0.9100 $\pm$ 0.0075 & 0.9325 $\pm$ 0.0081 \\
        VFL Baseline           & 0.7318 $\pm$ 0.0122 & 0.7123 $\pm$ 0.0131 & 0.7766 $\pm$ 0.0133 & 0.6960 $\pm$ 0.0114 & 0.7766 $\pm$ 0.0132 \\
        \textbf{HybridVFL}     & \textbf{0.8928 $\pm$ 0.0113} & \textbf{0.8732 $\pm$ 0.0101} & \textbf{0.9235 $\pm$ 0.0146} & \textbf{0.8880 $\pm$ 0.0127} & \textbf{0.9235 $\pm$ 0.0146}\\
        \bottomrule
        \end{tabular*}
        \end{strip}

    \subsection{Superiority of HybridVFL in the Federated Setting}
    
        The comparison of centralized models highlights the necessity of multimodal features. The Centralized Multimodal model achieved a balanced accuracy of 0.9325, significantly surpassing the Centralized Image-Only baseline at 0.9074. This gap validates the critical role of tabular metadata in effective skin lesion classification, supporting the core Vertical Federated Learning approach.
        The results also demonstrate the clear superiority of \textbf{HybridVFL} over the standard VFL baseline. 
        In the privacy-preserving configuration, \textbf{HybridVFL} achieves a balanced accuracy of 0.9235, an improvement of 14.69 percentage points over the 0.7766 baseline using simple concatenation. 
        This validates the central hypothesis that a semantically aware fusion architecture yields tangible performance gains in VFL. 
        These gains stem from the framework’s core components: client-side feature disentanglement and server-side Transformer-based contextual fusion, which together produce richer multimodal representations from distributed data.

    \subsection{The Cost of Privacy}
    
        The minimal cost of privacy for \textbf{HybridVFL} is low, evidenced by only a 0.9-point balanced accuracy gap (0.9235 vs. 0.9325) compared to the Centralized Multimodal model. This low gap is attributed to the framework's sophisticated architecture, utilizing client-side feature disentanglement and a server-side Cross-Modal Transformer. Furthermore, HybridVFL significantly outperforms the less robust Centralized Image-Only baseline by 1.61 percentage points (0.9235 vs. 0.9074). This outcome demonstrates the effectiveness of the advanced multimodal fusion in balancing performance with data decentralization and compliance with GDPR and HIPAA.

\section{Conclusion and Future Directions}

    In this work, we introduced HybridVFL, a Vertical Federated Learning framework designed to overcome the limitations of simple fusion in edge-based multimodal applications. By integrating client-side feature disentanglement with a server-side Cross-Modal Transformer, HybridVFL achieves a 14.69\% improvement in balanced accuracy over standard concatenation baselines on the HAM10000 dataset. This performance approaches that of centralised oracles while crucially retaining raw data on local clients, demonstrating the viability of sophisticated fusion strategies for privacy-preserving edge intelligence in healthcare.

    \subsection*{Future Directions}
    
        While effective, this study has limitations that define our future roadmap. First, our current evaluation is limited to a single dataset and two modalities; future work will expand to diverse clinical benchmarks and incorporate additional modalities such as genomic data. Second, while we analysed communication costs, we will explicitly quantify computational latency and energy consumption on varied edge hardware to ensure real-world feasibility. Finally, although HybridVFL prevents raw data exposure, exchanging embeddings can still be susceptible to inversion attacks~\cite{jin2021cafe}. We plan to integrate differential privacy or secure multi-party computation (MPC) to robustly defend against such adversarial threats in production deployments.

\section*{Code Availability}

    The code and experiment results are available for viewing at: 
    \href{https://github.com/mociatto/HybridVFL}{\texttt{https://github.com/mociatto/HybridVFL}}

\bibliographystyle{ACM-Reference-Format}
\bibliography{reference}

\end{document}